\documentclass[letterpaper, conference, 10pt,twocolumn ]{IEEEtran}
\pdfoutput=1
\let\chapter\section
\usepackage[ruled,vlined,linesnumbered]{algorithm2e}
\usepackage{graphicx}
\usepackage{amsfonts}
\usepackage{amsmath,amssymb,mathrsfs}
\usepackage{array}
\usepackage{flafter}
\usepackage{cite}
\usepackage{subfigure}
\usepackage{verbatim}
\usepackage{caption}
\usepackage{color}
\usepackage[numbers,sort&compress]{natbib}
\usepackage{bm}
\usepackage{flushend}

\newtheorem{assumption} {Assumption}
\newtheorem{proposition} {Proposition}

\newcommand{\pbold}[1]{\textbf{#1}: }
\newcommand{\pitalic}[1]{\textit{#1}: }

\IEEEoverridecommandlockouts

\begin{document}
%
\title{Motion Planning for Global Localization in Non-Gaussian Belief Spaces}

\author{\IEEEauthorblockN{Saurav Agarwal}
\and
\IEEEauthorblockN{Amirhossein Tamjidi}
\and
\IEEEauthorblockN{Suman Chakravorty}
\thanks{Saurav Agarwal (\texttt{sauravag@tamu.edu}), Amirhossein Tamjidi (\texttt{ahtamjidi@tamu.edu}) and Suman Chakravorty (\texttt{schakrav@tamu.edu}) are with the Department of Aerospace Engineering, Texas A\&M University, College Station, TX 77840.}}

\maketitle

\begin{abstract}
This paper presents a method for motion planning under uncertainty to deal with situations where ambiguous data associations result in a multimodal hypothesis on the robot state. In the global localization problem, sometimes referred to as the ``lost or kidnapped robot problem", given little to no a priori pose information, the localization algorithm should recover the correct pose of a mobile robot with respect to a global reference frame. We present a Receding Horizon approach, to plan actions that sequentially disambiguate a multimodal belief to achieve tight localization on the correct pose in finite time, i.e., converge to a unimodal belief. Experimental results are presented using a physical ground robot operating in an artificial maze-like environment. We demonstrate two runs wherein the robot is given no a priori information about its initial pose and the planner is tasked to localize the robot.
\end{abstract}

\IEEEpeerreviewmaketitle

\section{Introduction}
\begin{figure}[h!]
  \centering
 {\includegraphics[height=1.8in]{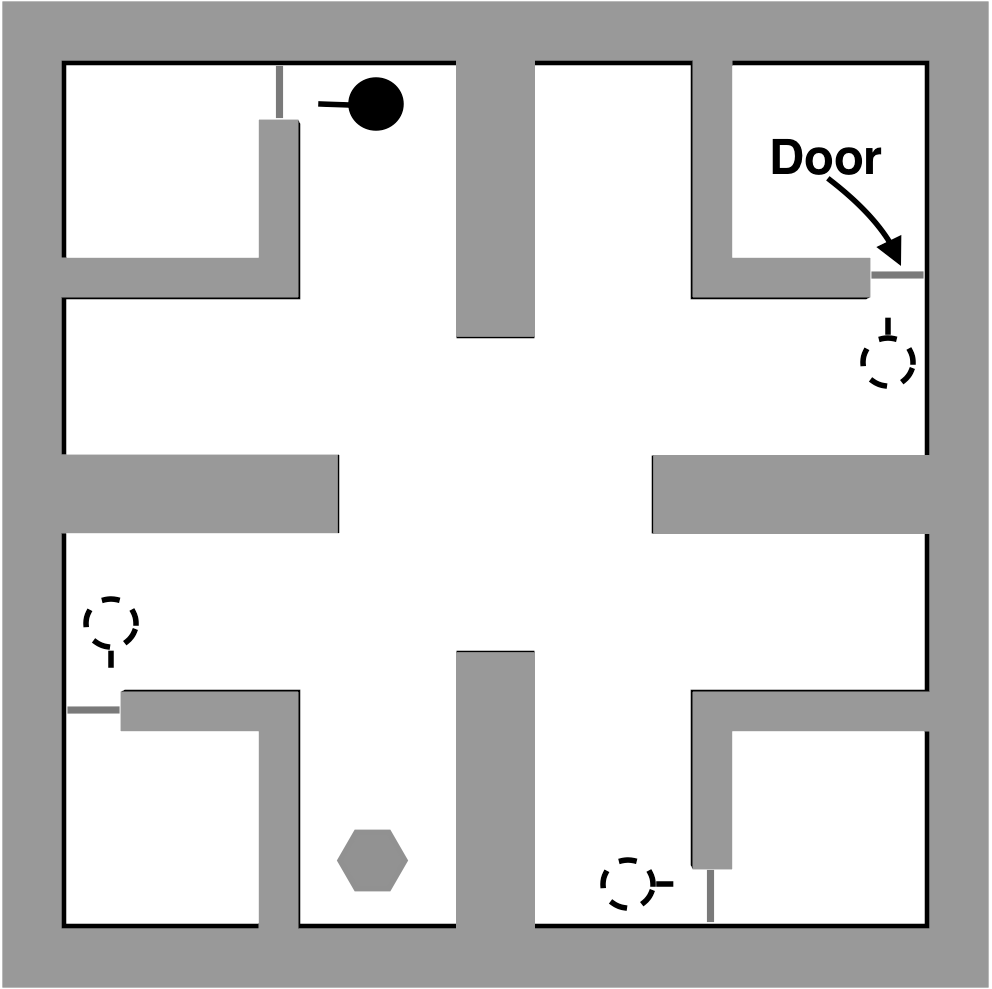}}
  \caption{A simple scenario depicting a multi-hypothesis localization problem, in a world with 4 rooms with identical doors. The true hypothesis is depicted by the solid disk, whereas the others are depicted by dashed disks. As the robot cannot distinguish between the doors, all hypotheses are equally likely.}
  \label{fig:multi-modal-example}
\end{figure}

In the domain of motion planning for mobile robots, situations may arise where data association between what is observed and the robot's map leads to a multimodal hypothesis on the state, for example a kidnapped robot with no a priori information or a mobile robot operating in a symmetric environment (see Fig. \ref{fig:multi-modal-example}). A large class of planning problems (e.g. visiting a fixed way point in space, driving through a narrow passage) require a well localized belief. State of the art belief space planning methods \cite{Prentice09,Bry11,chaudhari-ACC13,Berg10,kurniawati2012global,Ali14-IJRR,Platt10} rely on a Gaussian belief assumption to create solutions in the belief space. However, as discussed above, the Gaussian (unimodal) belief assumption may not always be a valid choice. 
This creates the requirement for a planner that can ``actively'' disambiguate a multimodal hypothesis. 

We represent a multimodal hypothesis with a Gaussian Mixture Model (GMM) and use an Extended Kalman filter (EKF) based Multi-Hypothesis Tracking (MHT) approach to propagate the belief. Our Multi-Modal Motion Planner (M3P) achieves disambiguation in a multimodal belief by first finding a neighboring location (referred to as target state) for each belief mode and then creating a candidate action to guide the belief mode to its target state such that these actions lead to information gathering behavior. The target states are chosen such that different modes of the robot's belief are expected to observe distinctive information, thus accepting or rejecting hypotheses in the belief.

\pbold{Contributions} The main contributions of this work can be summarized as follows:
\begin{enumerate}
	
	\item We develop a novel method for picking target states and creating candidate trajectories for a multimodal belief and choosing the best one, such that the maximum disambiguating information is observed which helps in rejecting incorrect hypotheses. 
	
	\item We prove that under certain realistic assumptions, through a process of iterative hypothesis elimination, our method can localize to the true robot pose.
	
	\item We demonstrate our method on a 2D navigation problem in which a robot begins in a kidnapped situation and recovers its pose.
	
\end{enumerate}


Albeit the scenario used to motivate the problem is the kidnapped robot situation, the method proposed is general, and can be extended to any planning situation where a multimodal belief arises in the robot state due to ambiguous data associations (a common practical issue in robot localization \cite{thrun2005probabilistic}). In the proceeding section, we present relevant related work, and discuss how our approach compares with them. In Section \ref{sec:problem} we state some preliminaries followed by the problem formulation. In Section \ref{sec:method} we present our method followed by experimental results in Section \ref{sec:experiments}.

\section{Related Work}\label{sec:relatedwork}

Recent work in sampling-based methods for belief space planning have shown promising results. The fundamental goal being to plan actions that minimize uncertainty such that a mobile robot can localize accurately to act safely and reliably. Methods such as \cite{Prentice09,Bry11,chaudhari-ACC13,Berg10,kurniawati2012global, Platt10} provide solutions that depend on the initial belief. Recent developments in \cite{Ali14-IJRR, Ali14-RolloutFIRM-ICRA} extend belief space planning to multi-query settings (cases where multiple planning requests are made sequentially) by creating a belief space variant of a Probabilistic RoadMap (PRM) \cite{Kavraki96}. We note that all the methods mentioned above rely on a Gaussian belief assumption. Additionally, the aforementioned methods assume that the data associations between observations and information sources (e.g., landmarks) are known and unambiguous. In contrast, this work does not assume that the data associations are unambiguous or that the belief is unimodal.
Another class of methods is the trajectory optimization approach which can be implanted in a Receding Horizon Control (RHC) framework for planning. A widely used approach in RHC-based control is to approximate the stochastic system with a deterministic one by substituting the random variables with their most-likely values \cite{Bertsekas07}. Methods such as \cite{Chakrav11-IRHC, Platt10, He11JAIR}, as well as the work in this paper assume the most-likely values for the unknown future observations in the planning stage.

Recent work in \cite{Platt11-isrr, platt-correctness-icra12} extends belief space planning to non-Gaussian beliefs. The authors investigate a grasping problem with a multimodal hypothesis on the gripper's state. Their method picks the most-likely hypothesis and a fixed number of samples from the belief distribution, then using an RHC approach, belief space trajectories are found that maximize the observation gap between the most-likely hypothesis and the drawn samples, which helps to accept or reject the most-likely hypothesis. The method in \cite{platt-wafr12-RHC} builds upon the work in \cite{Platt11-isrr} wherein the author transposes the non-convex trajectory planning problem in belief space to a convex problem. Compared to \cite{Platt11-isrr, platt-correctness-icra12, platt-wafr12-RHC}, our method is better suited to deal with more severe cases of non-Gaussian belief space planning such as the kidnapped robot scenario. Such scenarios may not be possible to address using the trajectory optimization based techniques of \cite{platt-wafr12-RHC, Platt11-isrr} in their current form, due to the difficulty of generating an initial feasible plan for the widely separated modes in the presence of obstacles (see Fig. \ref{fig:multi-modal-example} for an example of widely separated modes).

In the domain of global localization with a priori maps, \cite{dudek-whitesides} showed that finding the optimal (shortest) plan to re-localize a kidnapped robot with multiple hypothesis in a deterministic setting (no sensing or motion uncertainty) is NP-hard. At best a greedy localization strategy can be developed whose plan length is upper bounded by $(n-1)d$, where $n$ is the number of hypothesis and $d$ is the length of the optimal plan. Compared to \cite{dudek-whitesides}, we do not assume perfect sensing or actuation. 
In \cite{fox1998aml}, the authors develop an active localization method in a grid based scheme for a known map. Their planning method considers arbitrary targets in the robot's local coordinate frame as atomic actions (e.g., move 1m right and 4m forward). The optimal action is selected based on the path cost and the expected decrease in entropy at the target. Compared to \cite{fox1998aml}, our target selection methodology (Section \ref{subsec:offlinephase} and \ref{subsubsec:pick-target}) is active, i.e., M3P uses the a priori map information to select targets such that by visiting them, observation gap between belief modes is maximized resulting in successive disambiguation.

Successful application of the Gaussian mixture model to multi-hypothesis tracking for robot localization was shown in \cite{reuter2000, jensfelt-tro-2001, roumeliotis2000bayesian}. In \cite{jensfelt-tro-2001}, the authors present a greedy heuristic-based planning strategy to disambiguate a multimodal hypothesis along with an experimental demonstration for a kidnapped robot. The paper alludes to the fact that planning can be improved with a POMDP (Partially Observable Markov Decision Process) style approach. The method of \cite{hybrid-loc-ulivi} uses a hybrid approach in which a particle filter is used for hypothesis generation and an EKF is used to track each hypothesis, safe trajectories are planned by picking a point in the vicinity of obstacles to disambiguate the hypothesis. Compared to \cite{jensfelt-tro-2001, hybrid-loc-ulivi}, we present a planning approach that explicitly reasons about the belief evolution as a result of actions in the planning stage and picks an optimal policy from a set of candidates. We now proceed to formally describe our problem statement.

\section{Problem statement}\label{sec:problem}

Let $ x_{k} $, $ u_{k}$, and $ z_{k}$ represent the system state, control input, and observation at time step $ k $ respectively. Let $ \mathbb{X} $, $ \mathbb{U} $, and $ \mathbb{Z} $ denote the state, control, and observation spaces respectively. It should be noted that in our work, the state $x_k$ refers to the state of the mobile robot, i.e., we do not model the environment and obstacles in it as part of the state. The sequence of observations and actions are represented as $ z_{i:j}=\{z_{i},z_{i+1},\dots,z_{j}\} $ and $ u_{i:j}=\{u_{i},u_{i+1},\dots,u_{j} \}$ respectively. The non-linear state evolution model $ f $ and measurement model $ h $ are denoted as  $x_{k+1}=f(x_{k},u_{k},w_{k})$ and $z_{k}=h(x_{k},v_{k})$, where $w_{k} \sim \mathcal{N}(0,Q_k)$ and $ v_{k} \sim \mathcal{N}(0,R_k)$ are zero-mean Gaussian process and measurement noise, respectively.

The belief $b_k$ at time $t_k$ can be represented by a Gaussian Mixture Model (GMM) as a weighted linear summation over Gaussian densities. Let $w_{i,k}$, $\mu_{i,k}$ and $\Sigma_{i,k}$ be the weight, mean vector, and covariance matrix associated to the $i^{th}$ Gaussian $m_{i,k}$ respectively at time $t_k$, then

\begin{equation}\label{eq:gmm-linear-sum}
b_k = \sum_{i=1}^{M_k} w_{i,k} m_{i,k},  ~~ m_{i,k} \sim \mathcal{N}(\mu_{i,k}, \Sigma_{i,k}),
\end{equation}

where $M_k$ is the number of modes at time $t_k$. We state our problem as follows:

\textit{Given an a priori map, the system dynamics and observation models, construct a belief space planner $G(b_k)$ such that under the planner $G$, given any initial multimodal belief $b_0$, the belief state process evolves such that $M_{k} = 1$, for some finite time steps $k$.}\\

Note that in certain cases, the map may not allow actions that lead to hypothesis elimination such that the belief converges to a unimodal distribution (e.g., in a map with two identical closed rooms, if a robot is kidnapped and placed in either room, at best the robot can assign some probability to being in each room based on its observations). In such cases, M3P attempts to minimize $M_k$ (by design). Moreover, note that it is not possible to pre-compute what this minimum value of $M_k$ is without knowing the true hypothesis in the multimodal belief. 

\section{Methodology}\label{sec:method}

We begin by defining certain key concepts used in the M3P planner.

\pitalic{Uniqueness Graph} A graph $U_g$, whose nodes are states sampled from the collision free space and whose edges relate the similarity of information observed at the sampled locations.

\pitalic{Target State} A target state $v^{tt}_{i} \in U_g$ for mode $m_i$ is a node of the uniqueness graph which belongs to some neighborhood of radius $R$ of the mode's mean $\mu_{i}$ such that if each mode were to visit its target, the observations at the target would lead to disambiguation in the belief.

\pitalic{Candidate Policy}A candidate policy $\pi_i$ for mode $m_i$ is a local feedback controller that guides the mode to its target $v^{tt}_{i}$.

The M3P methodology has two phases, an offline phase in which we generate $U_g$ and an online phase in which we use the offline computations and plan in a receding horizon manner to disambiguate the belief.


\begin{figure*}
	\centering
	\subfigure[Candidate A leads to negative information for the mode in lower left corner. It expects to see the distinctive landmark which robot doesn't observe, and is thus rejected.]{\includegraphics[width=3.2in]{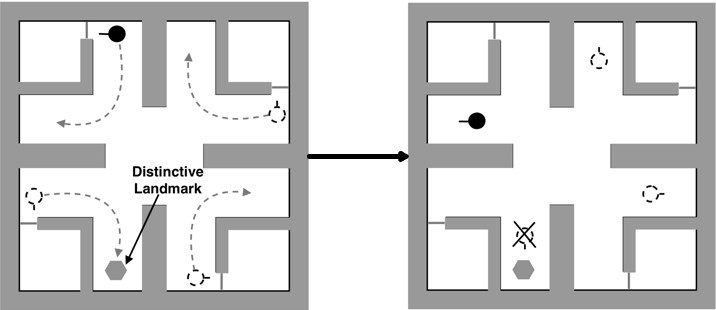}}
	\hspace{0.3in}
	\subfigure[Candidate B leads the true hypothesis to be confirmed as the robot sees the distinctive landmark.]{\includegraphics[width=3.2in]{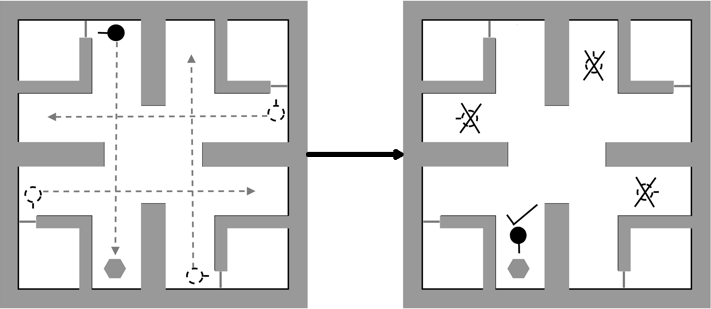}}
	\caption{Extending the example in Fig. \ref{fig:multi-modal-example}, we depict how M3P creates candidate trajectories and picks the optimal one. For clarity we show only two candidates A \& B and the effect of their execution. Candidate B results in complete dismabiguation and is clearly a better choice.}
	\label{fig:planner-rhc-method}
\end{figure*}

\subsection{Computing the Uniquenss Graph: Offline Phase}\label{subsec:offlinephase}

The uniqueness graph $U_g$ is constructed by uniformly sampling the configuration space and adding these samples as nodes of $U_g$. Once a node is added, we simulate the observation for the state represented by that node. Let $v_{\alpha}$ be one such node and $z^{v_{\alpha}}$ be the observation if the robot were to be in state $v_{\alpha}$. We add an edge $E_{\alpha \beta}$ (undirected) between two nodes $v_{\alpha}$ and $v_{\beta}$ if the simulated observations at both nodes are similar. Further, the edges are weighted and the weight is dependent on the degree of similarity in the information observed. For example, in our landmark based observation model, each landmark has a signature (appearance), thus if the state at $v_{\alpha}$ observes $z^{v_{\alpha}} = \{s_1,s_2,s_3\}$, i.e., the landmarks with signature $s_1,s_2$ and $s_3$ and at $v_{\beta}$ observes $z^{v_{\beta}} = \{s_1,s_2,s_4\}$, the edge $E_{\alpha \beta}$ would be given a weight of $2$ as there two similar landmarks observed ($z^{v_{\alpha}} \cap z^{v_{\beta}} = \{s_1,s_2\}$). A higher edge weight signifies that the states represented by the vertices of that edge are more likely to observe similar information. The lack of an edge between two nodes means that if a robot were to make an observation at those two states, it would see distinctly different information. The complexity for the construction of $U_g$ is $\mathcal{O}(n^2)$ (where $n$ is the number of samples) as each sample (node) is checked with every other for information overlap.

\pitalic{Issues} Due to its random nature, sampling may often occur in regions of low information density (e.g., regions where there are few or no landmarks). One can often circumvent this issue by increasing the number of samples. As $U_g$ is computed offline, the online performance is not significantly affected. Recent work in \cite{locawaresampling-iros15} suggests a localization aware sampling strategy which may be explored in future work.

\subsection{RHC based Planning: Online Phase}\label{subsec:onlinephase}
In a multimodal scenario, we claim that the best action to take is one that guides a robot without collision through a path that results in information gain such that a disambiguation occurs (one or more hypotheses are rejected, see Fig. \ref{fig:planner-rhc-method}). 
Algorithm \ref{alg:m3p} describes the online planning process. In step 3, the planner picks target states for each belief mode such that visiting a target can either prove or disprove the hypothesis. In step 4, the planner generates a set of candidate policies to drive each mode to its target. In step 5, the expected information gain for each policy is computed and we pick the best one, and in step 7, the multimodal belief is propagated according to the action and observations. We proceed to describe steps 3, 4, 5 and 7 of Algorithm \ref{alg:m3p} below.

\begin{algorithm}[h!]
	\caption{M3P: MultiModal Motion Planner} \label{alg:m3p}
	Input: $b$ \\
	
	\While{$b \neq \mathcal{N}(\mu, \Sigma)$}
	{
		$\{v^{tt}\} \gets$ Pick target states for belief modes (see Alg. \ref{alg:find-target-node});\\
		
		$\Pi \gets $ Generate candidate policies to connect each mode to its target;\\
		
		$\pi^* \gets $ Pick policy from $\Pi$ with maximum expected information gain; \\
		
		\ForAll{$u \in \pi^*$}
		{
			$b \gets $ Apply action $u$ and update belief (see Alg. \ref{alg:gmm-weight-update} for weight update calculation);\\
			
			\If{Change in number of modes or Expect a belief mode to violate constraints}
			{
				break; \\
			}
		}
	}
	
	\Return $b$;
\end{algorithm}

\subsubsection{Picking the target state for a mode}\label{subsubsec:pick-target}
Let us pick a mode $ m_{i,k} \sim \mathcal{N}(\mu_{i,k}, \Sigma_{i,k})$ from the belief. To find the target for $m_{i,k}$, we first choose the set of nodes $N_{i,k} \in U_g$ (Section \ref{subsec:offlinephase}) which belong to the neighborhood of the mean $\mu_{i,k}$ at time $t_k$. Then, we find the target node $v^{tt}_{i,k} \in N_{i,k}$ which observes information that is least similar in appearance to that observed by nodes in the neighborhood $N_{j,k}$ of mode $m_{j,k}$ where $j \neq i$. We are trying to solve the optimization problem,

\begin{equation}\label{eqn:findtargetnodes}
\{v^{tt}_{1,k}, v^{tt}_{2,k}, \dots ,v^{tt}_{M_k,k}\} = \operatorname*{arg\,min}_{v_{i,k} \in N_{i,k}} \bigcap^{M_k}_{i=1}z^{v_{i,k}},
\end{equation}

where $z^{v_{i,k}} = h(v_{i,k}, 0)$. Thus, Eq. \ref{eqn:findtargetnodes} solves for the set of target nodes, such that if each mode $m_{i,k}$ were to visit its target node $v^{tt}_{i,k}$, then we can prove or disprove the $i$-th mode. To solve Eq. \ref{eqn:findtargetnodes}, first we calculate the total weight of the outgoing edges from every node $v_{i,k} \in N_{i,k}$ to nodes in all other neighborhoods $N_{j,k}$ where $j \neq i$. The node which has the smallest outgoing edge weight, is the target candidate $v^{tt}_{i,k}$ for $m_{i,k}$ as the observation $z^{v^{tt}_{i,k}}$ would be least similar to the information observed in the neighborhood of all other modes $m_j$ where $j\neq i$. Algorithm \ref{alg:find-target-node} describes in detail the steps involved. 

\begin{algorithm}[h!]
	\caption{Finding the target for $i$-th mode} \label{alg:find-target-node}
	{
		Input: $b_k$, $i$ , $U_g$  \\ 
		Output: $v^{tt}_{i,k}$ \\ 
		
		\ForAll{$l \in [1,M_k]$}
		{
			$N_{l,k} \gets$ Find nodes in $U_g$ within neighborhood of $\mu_{l,k}$;\\
		}
		
		$minWeight \gets$ Arbitrarily large value; \\
		
		$v^{tt}_{i,k} \gets -1$; \\
		
		\ForAll{$ v \in N_{i,k}$ }
		{
			$w \gets 0$;\\
			\For{$ N_{j,k} \in \{N_{1,k},\dots, N_{M_k,k}\} \setminus N_{i,k}$}
			{
				
				\ForAll{$e \in$ Edges connected to $v$}
				{
					\ForAll{$p \in N_{j,k} $}
					{
						\If{$p$ is a target of edge $e$}
						{
							$w \gets w + \mathtt{edgeWeight}(e)$; \\
						}

					}
				}
				
			}
			\If{$w < minWeight$}
			{
				$minWeight \gets w$; \\
				$v^{tt}_{i,k} \gets v$;\\
			}
		}
		
		\Return $v^{tt}_{i,k}$; \\
	}
\end{algorithm}

\subsubsection{Generating candidate policies for belief modes}\label{subsubsec:genpolicies}
Once we have picked the targets corresponding to each mode, we need to find the control action that can take a mode from its current state to the target state. We generate the candidate trajectory that takes each mode to its target using the RRT* planner \cite{Karaman.Frazzoli:IJRR11}. Once an open loop trajectory is computed, we generate a local policy $\pi_i$ (feedback controller) for the $i$-th mode, which drives the $i$-th mode along this trajectory. Let $\Pi$ be the set of all such policies for the different modes.

\subsubsection{Picking the Optimal Policy}\label{subsubsec:optimalpolicy}
Once we have generated the set $\Pi$ of candidate policies. We need to evaluate the expected information gain $\Delta I_i$ for each policy $\pi_i$ and pick the optimal policy $\pi^*$ that maximizes this information gain. We model this information gain as the discrete change in the number of modes. To compute the expected change in the number of belief modes, we simulate the most-likely belief trajectory, i.e., approximating noisy observations and actions with their most-likely values. We know that one policy may or may not be good for all the modes, i.e., a policy based on one mode may lead to collision for the other modes. Therefore, we need a way of penalizing a candidate policy during the planning stage if it results in collision. We introduce a penalty $c_{fail} / T$ into the information gain where $c_{fail}$ is a fixed value and $T$ is the time step at which the collision takes place during the simulation of belief trajectory under some policy. Thus, policies which result in a collision much further down are penalized less compared to policies that result in immediate collision. The steps to calculate the expected information gain for a candidate policy $\pi_i \in \Pi$ are as follows:

\begin{enumerate}
	\item For every belief mode $m_{j,k} \in b_k$. 
	\begin{enumerate}
	 \item Assume that robot is at  $m_{j,k}$.
	 \item Simulate $\pi_i$ and propagate all the modes. 
	 \item Compute information gain $\Delta I_{i,m_{j,k}}$ (reduction in number of modes while factoring in collision cost) for $\pi_i$.
	\end{enumerate}
	\item Compute the weighted information gain $\Delta I_{i} = \sum^{M_k}_{j=1} w_{j,k}\Delta I_{i,m_{j,k}}$.
\end{enumerate}

After computing the expected information gain for each policy, we pick the gain maximizing policy. The computational complexity of picking the optimal policy is $\mathcal{O}(M_k^3 L_{max})$ (where $M_k$ is the number of belief modes and $L_{max}$ is the maximum candidate trajectory length). This is due to the fact that each policy is simulated for each mode for the length of policy, where at every step of policy execution, there are $M_k$ filter updates. Figure \ref{fig:planner-rhc-method} depicts the process of picking the optimal candidate trajectory in a simple scenario. 

\subsubsection{Belief Propagation Using GMM}\label{subsec:gmm-weight}

We first discuss our decision to use EKF based multi-hypothesis tracking over a particle filtering approach. In practical localization problems, a relatively small number of Gaussian hypothesis are sufficient for maintaining the posterior over the robot state, secondly the filtering complexity grows linearly in the number of hypothesis and finally due to the complexity of our re-planning step ($\mathcal{O}(M_k^3 L_{max})$), the number of samples required for a particle filter would render re-planning computationally inefficient. Now, we proceed to describe the weight update step which determines how likely each mode is in the belief. In a standard implementation, the weights $w_{i,k}$'s are updated based on the measurement likelihood function as shown in Eq. \ref{eq:weight-update}.

\begin{equation}\label{eq:weight-update}
 w_{i,k+1} = w_{i,k} e^{-\frac{1}{2}D_{i,k+1}^{2}}
\end{equation}

where $D_{i,k+1}$ is the Mahalanobis distance between the sensor observation and most-likely observation for mode $m_i$ such that

\begin{equation}\label{eq:mahalanobis-distance}
D_{i,k+1}^{2} =  (z_{k+1}-h(\mu_{i,k+1},0))^T R_k^{-1} (z_{k+1}-h(\mu_{i,k+1},0)).
\end{equation}

The weights are normalized such that $\sum_{i=1}^{M_k} w_{i,k+1}=1$. A known issue with EKF-based MHT is that it is unable to process negative information \cite{thrun2005probabilistic}. Negative information refers to the lack of information which one may expect to see and can certainly help in disproving a hypothesis. We now proceed to describe how we factor in negative information in the weight update.

\pitalic{Factoring Negative Information} Depending on the state of the robot, individual hypotheses and data association results, we might have several cases. We discuss this issue in the context of a landmark based measurement model. At time $t_{k+1}$, let $n_{z_{k+1}}$ be the number of landmarks observed by the robot and $n_{z^p_{i,k+1}}$ be the number of landmarks that we predict to see for $m_i$ where $z^p_{i,k+1} = h(\mu_{i,k+1},0)$ is the predicted observation. Then $n_{z_{k+1}} = n_{z^p_{i,k+1}}$ means that the $i$-th mode expected to see as many landmarks as the robot observed; $n_{z_{k+1}} > n_{z^p_{i,k+1}}$ implies the robot observes more landmarks than predicted for the mode; $n_{z_{k+1}} < n_{z^p_{i,k+1}}$ implies the robot observes less landmarks than predicted for the mode. Also, we can have the number of data associations to be less than the number of predicted or measured observations or both. This means that we may not be able to make a unique association between each predicted and some observed landmark. At time $t_{k+1}$, we estimate the Mahalanobis distance $D_{i,k+1}$ (Eq. \ref{eq:mahalanobis-distance}) for mode $m_i$ between the predicted and observed landmarks that are matched by the data association module and update weight according to Eq. \ref{eq:weight-update}. Then we multiply the updated weight by a factor $\gamma$, which models the effect of duration $\beta_{i,k+1}$ for which the robot observes different landmarks than the $i$-th mode's prediction; and the discrepancy $\alpha$ in the number of data associations. When a belief mode is initialized, we set $\beta_{i,0} = 0$. The weight update procedure is described in Algorithm \ref{alg:gmm-weight-update}. After each weight update step, we remove modes with negligible contribution to the belief, i.e., when $w_{i,k+1} \leq \delta_w$ where $\delta_w$ is a user defined parameter for minimum weight threshold \footnote{In practical MHT applications, highly similar modes can often be merged to reduce complexity. We refer the reader to \cite{thrun2005probabilistic,jensfelt-tro-2001} for a thorough explanation. From a planning perspective, when two modes are merged, policy execution is halted, and re-planning is triggered. M3P computes candidate policies for the new set of belief modes to disambiguate them.}.


\begin{algorithm}[h!]
\caption{GMM Weight Update} \label{alg:gmm-weight-update}
Input: $w_{i,k}, \mu_{i,k+1}, \beta_{i,k}, \delta t$ \\
Output: $w_{i,k+1}, \beta_{i,k+1}$ \\

$z_{k+1}, n_{z_{k+1}} \gets $  Get sensor observations;\\


$z^p_{i,k+1}, n_{z^p_{i,k+1}} \gets $  Get predicted observations for $\mu_{i,k+1}$;\\


$n_{z_{k+1} \cap z^p_{i,k+1}} \gets$ Do data association;\\


$w'_{i,k+1} \gets$ Update and normalize weight according to likelihood function; \\

$ \gamma \gets 1$;\\
\If{ $n_{z^p_{i,k+1}} \neq n_{z_{k+1}} ~ \text{or} ~ n_{z^p_{i,k+1}} \neq n_{z_{k+1} \cap z^p_{i,k+1}} $}
{

  $\alpha \gets max( 1 + n_{z_{k+1}} - n_{z_{k+1} \cap z^p_{i,k+1}} , 1 + n_{z^p_{i,k+1}} - n_{z_{k+1} \cap z^p_{i,k+1}})$;\\ 
  $ \beta_{i,k+1} \gets \beta_{i,k} + \delta t $;\\
  $ \gamma \gets e^{-\alpha \beta_{i,k+1} 10^{-4}}$;\\
}

\Else
{
  $ \beta_{i,k+1} \gets 0$; \\
}

$w_{i,k+1}  \gets w'_{i,k+1} \gamma $;\\

\Return $w_{i,k+1}, \beta_{i,k+1}$;
\end{algorithm}

\subsection{Analysis}\label{subsec:analysis}

In this section, we show that the receding horizon planner M3P  will guarantee that an initial multimodal belief is driven into a unimodal belief in finite time. First, we make the following assumptions:
\begin{assumption}
Given a multimodal belief $b_k$, for every mode $m_{i,k}$, the environment allows for the existence of some target state $v^{tt}_{i,k}$ and some homotopy class of paths through which the robot can visit $v^{tt}_{i,k}$, such that if the robot was in mode $m_{i,k}$, it could confirm that $m_{i}$ is the true hypothesis. 
\end{assumption}
\begin{assumption}
The map does not change during the execution of the planner.
\end{assumption}

\begin{proposition}
Under Assumptions 1 and 2, given any initial multimodal belief $b_0 = \sum_i w_{i,0}m_{i,0}$, the receding horizon planner M3P drives the belief process into a unimodal belief $b_T = m_T \approx \mathcal{N}(\mu_T, \Sigma_T)$ in some finite time $T$.
\end{proposition}
\begin{IEEEproof}
Suppose that the robot is at the initial belief $b_0$. Suppose we choose the plan $G_{i^*}$, i.e., candidate policy for mode $i^*$, that results in the most information gain as required by the M3P planner. The plan $G_{i^*}$ can be applied to all the modes at least for some finite period of time in the future,  since if it cannot be applied, then we immediately know that the robot is not at mode $i^*$ and thus, there is a disambiguation whereby mode $i^*$ is discarded. Once the plan $G_{i^*}$ is executed, there are only 2 possibilities:\\
1) The robot is able to execute the entire plan $G_{i^*}$ till the end, or
2) the plan becomes unfeasible at some point of its execution. \\
In case 1, due to Assumption 1, we will know for sure that the robot was at mode $i^*$ and the belief collapses into a unimodal belief thereby proving the result. In case 2, due to Assumption 2, we know that the robot could not have started at mode $i^*$ and thus, the number of modes is reduced by at least one. After this  disambiguation, we restart the process as before and we are assured that at least one of the modes is going to be disambiguated and so on. Thus, it follows given that we had a finite number of modes to start with, the belief eventually converges to a unimodal belief. Further, since each of the disambiguation epochs takes finite time, a finite number of such epochs also takes a finite time, thereby proving the result.
\end{IEEEproof}

\pbold{Remarks} The above result shows that the M3P algorithm will stabilize the belief process to a unimodal belief under assumptions 1 and 2. In the case that assumption 1 is violated we are either (i) unable to find a target which allows the robot to observe distinctive information (e.g., trivial case of a robot operating in a world with identical infinite corridors) or (ii) we may find such a target but the environment geometry does not allow for any path to visit it (e.g., robot stuck in one of many identical rooms and the doors are closed) or (iii) all homotopy class of paths to visit a target state pass through long regions with no information such that the uncertainty on each mode grows sufficiently high that we cannot make data associations at the target location to disambiguate the multimodal belief. Violations (i) and (ii) refer to degenerate cases that rarely occur in practical motion planning problems. Violation (iii) is currently beyond the scope of this paper and presents an important direction for future research.
Assumption 2 (static world) is common in localization literature, though it may be violated in certain practical scenarios. In such cases, if the map is not changing rapidly, one may use sensory observations to incorporate new constraints into the map and trigger replanning.

\section{Experimental Results}\label{sec:experiments}


We present experimental results for a non-holonomic ground robot. The experiments represent two motion planning scenarios wherein the robot is placed randomly at a location in an environment which is identical to other locations in appearance. Thus the initial belief is multimodal, the goal of the experiment is to use the non-Gaussian planner M3P described in Section \ref{sec:method} to localize to a unimodal belief. We first describe the system setup to motivate the experiment followed by the results.

\begin{figure}[h!]
  \centering
 {\includegraphics[width=1.5in]{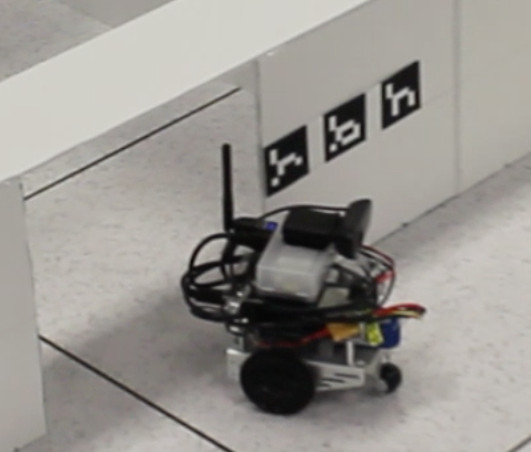}}
  \caption{The robot exiting a room in the maze. The robot has an $11.5$ cm wheelbase and measures $18$ cm and $15$ cm in height and length respectively.}
  \label{fig:robot}
\end{figure}

\subsection{System Description}

We used a low-cost Arduino based differential drive robot as shown in Fig. \ref{fig:robot}. The robot is equipped with an Odroid U3 computer running ROS on Ubuntu 14.04 and an off-the-shelf Logitech C-310 webcam for sensing. The onboard computer uses a wifi link to communicate with the ground control station. We use an off-the-shelf laptop running Ubuntu 14.04 as the ground control station. The ground station runs the planner and image processing algorithms while communicating with the robot via wifi.

\pitalic{Motion Model}The kinematics of the robot are represented by a unicycle.

\begin{align}\label{eq:unicycle-motion-model}
\!\!\!\!\!x_{k+1}& \!=\! f(x_k,u_k,w_k) \!=\!
\left(\!
  \begin{array}{c}
    \mathsf{x}_{k}+(V_k + n_v)\delta t\cos\theta_k \\
    \mathsf{y}_{k}+(V_k + n_v)\delta t\sin\theta_k \\
    \mathsf{\theta}_{k}+(\omega_k + n_{\omega})\delta t
  \end{array}\!\right)\!,
\end{align}
where $ x_k = (\mathsf{x}_k, \mathsf{y}_k, \mathsf{\theta}_k)^T $ describes the robot state (position and yaw angle). $ u_k = (V_k,\omega_k)^T $ is the control vector consisting of linear velocity $ V_k $ and angular velocity $ \omega_k $. We denote the process noise vector by $ w_k=(n_v,n_{\omega})^T\sim\mathcal{N}(0,\mathbf{Q}_k) $.


\pitalic{Observation Model} Our observation model is based on passive visual beacons/landmarks which can be observed by a monocular camera to measure their relative range, bearing and an associated ID tag. Let the location of the $i$-th landmark be denoted by ${^i}\mathbf{L}$. The displacement vector ${^i}\mathbf{d}$ from the robot to ${^i}\mathbf{L}$ is given by ${^i}\mathbf{d}=[{^i}d_{x}, {^i}d_{y}]^T:={^i}\mathbf{L}-\mathbf{p}$, where $\mathbf{p}=[\mathsf{x},\mathsf{y}]^T$ is the position of the robot. Therefore, the observation ${^i}z$ of the $i$-th landmark can be modeled as,

\begin{align}
{^i}z={^i}h(x,{^i}v)=[\|{^i}\mathbf{d}\|,\text{atan2}({^i}d_{y},{^i}d_{x})-\theta]^T+{^i}v,
\end{align}

The observation noise is zero-mean Gaussian  such that $ {^i}v\sim\mathcal {N}(\mathbf{0},{^i}\mathbf{R}) $ where ${^i}\mathbf{R}=\text{diag}((\eta_r\|{^i}\mathbf{d}\|+\sigma^r_b)^2,(\eta_{\theta}\|{^i}\mathbf{d}\|+\sigma^{\theta}_b)^2)$. 
The quality of sensor reading decreases as the robot gets farther from the landmarks. 
The parameters $\eta_r$ and $\eta_{\theta}$ determine this dependency, and $\sigma_b^r$ and $\sigma_b^{\theta}$ are the bias standard deviations. 


\begin{table}
\begin{center}
    \begin{tabular}{| l | l |}
    \hline
    \textbf{Parameter} & \textbf{Value} \\
    \hline 
    $\delta_w$ & $0.01$ \\
    \hline
    $c_{fail}$ & $10^6$ \\
    \hline
    RHC horizon & $60$ secs \\
    \hline
    \end{tabular}
     \caption{List of experiment parameters.}
     \end{center}
\end{table}

\begin{figure}[h!]
	\centering
	{\includegraphics[width=3in]{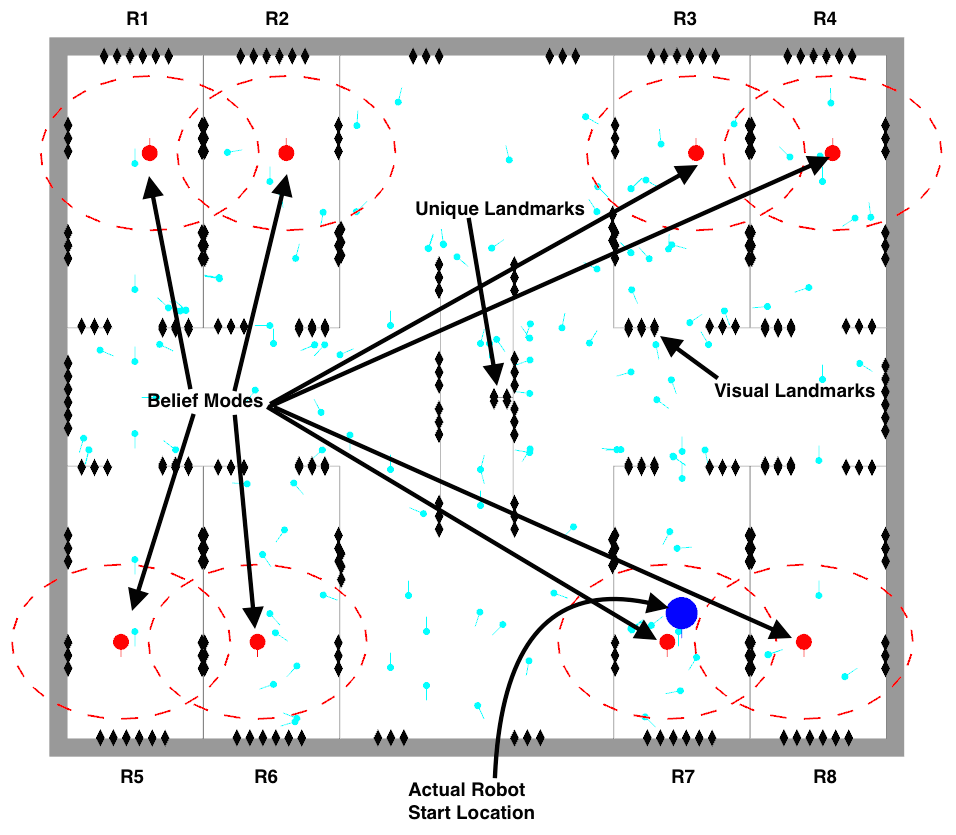}}
	\caption{The environment and belief at the start of first run. The robot is placed in room R7 (blue disk). The initial sampling leads to 8 belief modes, one in each room. The black diamonds mark the locations of augmented reality markers in the environment. The unique landmarks are placed inside the narrow passage, such that if the robot enters the passage from either side, it sees distinctive information.}
	\label{fig:experiment-run1-env}
\end{figure}

\subsection{Scenario}

We constructed a symmetrical maze that has 8 identical rooms (R1-8) as shown in Fig. \ref{fig:experiment-run1-env}. Augmented reality (AR) markers were placed on the walls which act as the landmarks detected by the vision-based sensing system on the robot \cite{Aruco2014}. Each marker has a signature (an ID number), thus when the robot sees a landmark, it can measure the range, bearing as well as its ID number. To create ambiguity in the data association, we placed multiple AR markers with the same ID number in different parts of the environment. For example, one of the symmetries in our experiment is the inside of each room. Each room in the maze appears identical to the robot as markers with the same appearance are placed on each room's walls with an identical layout. Thus, if the robot is placed in a location with markers similar to another part of the environment, the data associations lead the robot to believe it could be in one of these many locations, which leads to a multimodal belief on the state. We also place four unique landmarks in a narrow passage in the center of maze as marked in Fig. \ref{fig:experiment-run1-env}. To successfully localize, the robot must visit this location in order to converge to its true belief.

\begin{figure*}
	\centering
	\subfigure[The planner visualization showing the candidate trajectories (green). The top right image shows the view from the onboard camera, with the detected marker information overlaid. The bottom-right image shows the top-view of the maze in which the robot is run.]{\includegraphics[width=3.2in]{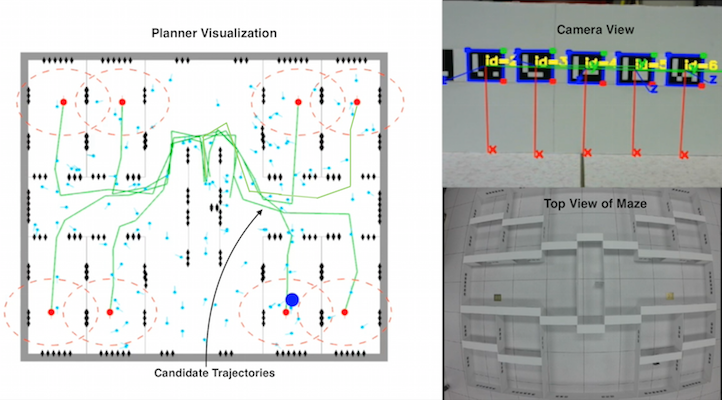}\label{fig:run1-phase2}}
	\hspace{0.1in}
	\subfigure[The robot observes landmark ID 55 on the door of the opposite room causing the weights of modes $m_1, m_3, m_4,m_5,m_6,m_8$ to gradually decrease which leads to these modes being removed from the belief.]{\includegraphics[width=3.2in]{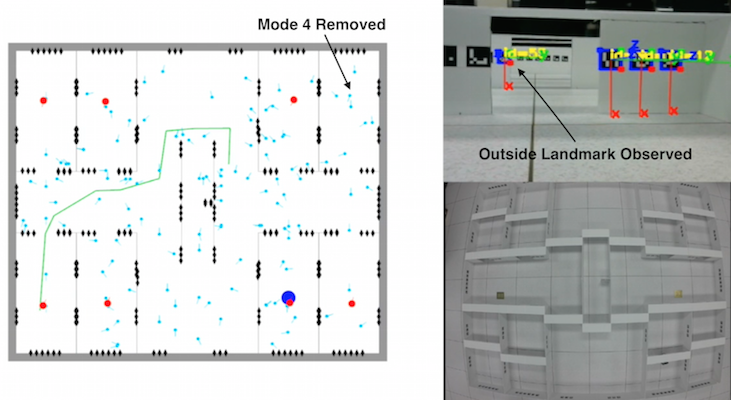}\label{fig:run1-phase2-2}}
	\vspace{0.2in}
	\subfigure[The robot has exited the room and is looking at the outside wall of the narrow passage. The two modes $m_2$ and $m_7$ are symmetrically located in the map, due to the information in the map that is observed by the robot.]{\includegraphics[width=3.2in]{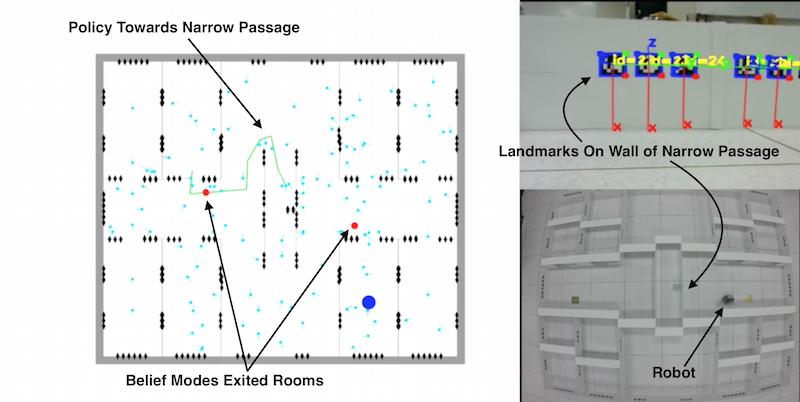}\label{fig:run1-phase3}}
	\hspace{0.1in}
	\subfigure[The belief mode has converged to the the true belief as the robot enters the narrow passage and observes the unique landmark (ID 39).]{\includegraphics[width=3.2in]{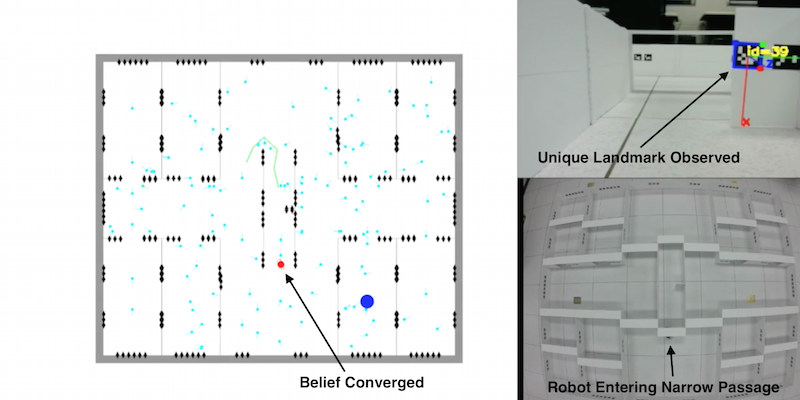}\label{fig:run1-phase4}}  
	\caption{Snapshots of the experiment at different times.}
	\label{fig:exprun}
\end{figure*}

The robot is initially placed in room R7 and is not given any prior information of its state. We assume a known map. To estimate $b_0$, we uniformly sample the configuration space and set these samples as the means $\mu_{i,k}$ of the modes of the Gaussian mixture components and assign identical covariance and uniform weight to each mode. After this, the robot remains stationary and the sensory measurements are used to update the belief state and remove the unlikely modes. Unlikely modes whose weight falls below a predesignated threshold $\delta_w$ are rejected. This process of elimination continues until we converge to a fixed number of modes. Figure \ref{fig:experiment-run1-env} shows the initial belief. The robot plans its first set of candidate actions as shown in Fig. \ref{fig:run1-phase2}. After the candidates are evaluated, the policy based on mode $m_5$ in room R5 is chosen and executed. As the robot turns, it sees a landmark on the wall outside the room (shown in Fig. \ref{fig:run1-phase2-2}). This causes mode $m_4$ to be deleted. Immediately, replanning is triggered and a new set of candidate trajectories is created. In successive steps, we see that first modes $m_3$ and $m_5$ are deleted and then after the next two replanning steps, modes $m_8$, $m_1$ and $m_6$ are deleted. We notice that the robot does not move till only the 2 most-likely modes are remaining. The reason for this is that seeing the marker on the outside wall has the effect of successively lowering the weights of the unlikely modes due to filtering. As the mode weights fall below the threshold, they are deleted, which triggers the replanning condition. Once the belief has converged to the two most-likely modes $m_2, m_7$ (as expected by the symmetry) a new set of candidate policies is created and the policy based on mode $m_2$ is chosen. This policy leads the modes out of the rooms, and towards the narrow passage. Figure \ref{fig:run1-phase3} shows both belief modes executing the policy based on mode $m_2$. While executing this policy, replanning is triggered as the robot exceeds maximum horizon for policy execution. The final policy drives the robot into the narrow passage and the unique landmarks are observed (Fig. \ref{fig:run1-phase4}) which leads the belief to converge to the true belief. 

\textit{Due to paucity of space we only present one experiment here, a supplementary video is provided that clearly depicts every stage of both our experiments}.

\subsection{Discussion} 
Our approach results in an intuitive behavior which guides the robot to seek disambiguating information such that it can sequentially reject the incorrect hypothesis about its state. The candidate trajectories are regenerated every time a Gaussian mode is rejected or a constraint violation is foreseen. The time to re-plan reduces drastically as the number of modes reduce. Thus, the first few actions are the hardest which is to be expected as we start off with a large number of hypotheses. Finally, the planner is able to localize the robot safely. In \cite{fox1998aml}, the authors showed that random motion is inefficient and generally incapable of localizing a robot within reasonable time horizons especially in cases with symmetry (e.g., office environments with long corridors and similar rooms). In \cite{jensfelt-tro-2001} the authors consider the robot localized when one of the modes gets a weight $\geq 0.8$, in contrast our approach is more conservative in that we only consider the robot localized when a mode has weight $\geq 0.99$. We can afford to be more conservative as our localization strategy actively seeks disambiguating information using prior map knowledge as opposed to a heuristic based strategy. While our experiment acts as a proof of concept, there are certain phenomenon such as cases where the belief modes split into child modes, or dynamic environments which were not covered and will be addressed in future work.

\section{Conclusion}
In this work, we studied the problem of motion planning for a mobile robot when the underlying belief state is non-Gaussian in nature. Our main contribution in this work is a planner M3P that generates a sequentially disambiguating policy, which leads the belief to converge to a unimodal Gaussian. We are able to show in practice that the robot is able to recover from a kidnapped state and localize in environments that present ambiguous data associations. Compared to previous experimental work, we take an active approach to candidate policy generation and selection. However, a key limitation arises in our policy selection step due to algorithmic complexity. This needs to be addressed in future work such that the algorithm can scale well to larger maps which may result in a greater number of hypotheses. 
In future work, experiments will be extended to full-scale problems (e.g., office like environments where symmetries are known to cause ambiguity in the belief) and more drastic localization failures (e.g., when a well-localized robot is kidnapped to a random location, the a priori information tends to mislead the robot). Finally, there may be tasks which are feasible with a multimodal distribution on the belief. Such cases present an interesting area for future motion planning research.
\small{
\section{Acknowledgments}
We would like to thank Shiva Gopalan for helping us in programming the robot and assembling the maze for the experimental work in this paper.
}
\newpage
\bibliographystyle{plainnat}

\bibliography{references}

\end{document}